\documentclass[pdflatex,sn-mathphys-num]{sn-jnl}

\usepackage{graphicx}%
\usepackage{multirow}%
\usepackage{amsmath,amssymb,amsfonts}%
\usepackage{amsthm}%
\usepackage{mathrsfs}%
\usepackage[title]{appendix}%
\usepackage{xcolor}%
\usepackage{textcomp}%
\usepackage{manyfoot}%
\usepackage{booktabs}%
\usepackage{algorithm}%
\usepackage{algorithmicx}%
\usepackage{algpseudocode}%
\usepackage{listings}%

\theoremstyle{thmstyleone}%
\theoremstyle{thmstyletwo}%

\theoremstyle{thmstylethree}%

\begin{document}

\title[Article Title]{Structural Positional Encoding for knowledge integration in transformer-based  medical  process monitoring}


\author*[1]{\fnm{Christopher} \sur{Irwin}}\email{christopher.irwin@uniupo.it}

\author[1]{\fnm{Marco} \sur{Dossena}}\email{marco.dossena@uniupo.it}

\author[1,2]{\fnm{Giorgio} \sur{Leonardi}}\email{giorgio.leonardi@uniupo.it}

\author[1,2]{\fnm{Stefania} \sur{Montani}}\email{stefania.montani@uniupo.it}

\affil*[1]{\orgdiv{DISIT, Computer Science Institute}, \orgname{University of Piemonte Orientale}, \orgaddress{\city{Alessandria}, \country{Italy}}}

\affil[2]{\orgdiv{Laboratorio Integrato di Intelligenza Artificiale e Informatica in Medicina DAIRI}, \orgname{Azienda Ospedaliera SS. Antonio e Biagio e Cesare Arrigo}, \orgaddress{\city{ Alessandria}, \country{Italy}}}


\abstract{Predictive process monitoring is a process mining task aimed at forecasting information about a running process trace, such as the most correct next activity to be executed.  In medical domains, predictive process monitoring can provide valuable decision support in atypical and nontrivial situations. Decision support and quality assessment in medicine cannot ignore domain knowledge, in order to be grounded on all the available information (which is not limited to data) and to be really acceptable by end users.

In this paper, we propose a predictive process monitoring approach relying on the use of a {\em transformer}, a deep learning architecture based on the attention mechanism. A major contribution of our work lies in the incorporation of ontological domain-specific knowledge, carried out through a graph positional encoding technique. 
The paper  presents and discusses the encouraging experimental result we are collecting  in the domain of stroke management.}




\maketitle

\newpage

\section{Introduction}
\label{intro}

The diffusion of advanced medical information systems is progressively enabling the automatic collection of patients' {\em traces}, i.e., the sequences of activities executed on patients during the care procedures implemented at a hospital organization \cite{Reichert:2012}.
Patients' traces are a valuable source of information for several analyses and investigations, within the field of process mining \cite{aalst:book:16}. Among the various available process mining techniques, predictive process monitoring \cite{Maggi:2014,Teinemaa:2019} is of particular interest. 
Predictive process monitoring aims at forecasting relevant information about a running
process trace; specifically, it exploits the already logged traces  to make
predictions about the running trace completion, such as suggesting the next activity to be executed, or estimating the remaining time/cost/resources required to completion itself.

In the medical field, predictive process monitoring can support better time and resource allocation; moreover, and most importantly, it can support decision-making in nontrivial cases. Indeed, despite the availability of clinical guidelines, it is worth noting that guidelines are ideal processes, designed for ideal patients, and meant to be applied in an ideal setting, where all the needed resources are always available \cite{Xu:2020}. In reality, this is often not the case: local resource constraints may prevent from the (timely) execution of the scheduled guidelines activities. Moreover, real patients may be atypical, for instance, due to the presence of comorbidities or rare disease variants. Finally, physicians may lack the necessary background knowledge to correctly interpret and apply guidelines in complex cases. Predictive process monitoring, by suggesting the next activity to be executed, can thus provide very helpful advice in these nontrivial situations.

Stemming from the considerations discussed above, in this paper we propose a predictive process monitoring approach for medical process traces; according to the most recent literature advances, we rely on the exploitation of a {\em Transformer}, a deep learning architecture based on the self-attention mechanism \cite{vaswani2017attention}, which has already demonstrated its utility in modeling similar tasks, as exemplified in \cite{processtransformer}.  A major contribution of our work lies in the introduction of a technique that incorporates domain-specific knowledge using an ontology. This augmentation is carried out through a graph positional encoding technique, enhancing the accuracy of our model.

While the approach is general, we were motivated by a specific application domain, that we selected for our experiments, namely the one of stroke management. Indeed, in this field, predictive process monitoring can provide a valuable contribution to medical decision-making. To mention a few examples, it can help in the choice of the correct thrombolytic drug, by considering how indications and contraindications of the available medications were balanced in past traces \cite{Xu:2020}, or it can help in deciding whether to confirm the presence of subarachnoid hemorrhage by taking into account pros and cons of a lumbar puncture \cite{eswa16}, always referring to already completed traces.
Indeed, the first experiments we performed, which are described in the following, have shown encouraging results.

The paper is organized as follows: Section \ref{related} introduces some related work. Section \ref{methods} presents the deep learning architecture we exploit. Section \ref{experiments} illustrates our experimental setting and results, while section \ref{conclu} is devoted to conclusions.

\section{Related work}
\label{related}

Predictive process monitoring has been afforded by different types of  machine learning techniques, such as Transition Systems \cite{Le2012}, Hidden Markov Models \cite{Lakshmanan2015}  and
Support Vector Machines \cite{Cabanillas:2014}.
In more recent approaches, however, deep learning architectures are being applied, and nowadays represent the state of the art in the field.  

Different deep learning architectures have been proposed to deal with the task of next activity prediction. 
The approach in \cite{Mehdiyev}, for instance, uses Autoencoders \cite{Hinton}.
In predictive monitoring, Autoencoders can successfully reduce the number of attributes of the input activities; however, they are unable to treat long (temporal) dependencies between activities of a trace.
Convolutional Neural Networks (CNNs) \cite{Alom} have been applied in \cite{Appice}, where sequential data in process traces are treated as a  one-dimensional grid. 

Due to the sequential nature of traces, however, Recurrent Neural Networks (RNNs) \cite{Pascanu} probably represent the most natural solution. In fact, RNNs can capture longer dependencies between
the activities of a trace, while in a CNN an activity
only depends on the k most recent activities, where k is the
size of the kernel.
Within RNNs, Long-Short Term Memory (LSTM) networks \cite {Hochreiter} represent a particularly performant approach. Indeed, LSTM can potentially learn the complex dynamics within the temporal ordering of input sequences; therefore, they are well suited to manage the sequential data of process activity logs. They can also manage long-distance dependencies between activities, since they implement a long-term memory where the information flows from cell to cell with minimal variations, keeping certain aspects constant during the processing of all inputs. The works in \cite{Evermann,Tax,Camargo,lstm-rev}, for instance, rely on LSTMs to predict the next activity of a running case.

Similarly to LSTMs, Gated Recurrent Networks (GRUs) \cite{cho2014learning} also create paths through time that allow the gradients to flow deeper in the sequence than in basic RNNs; with respect to LSTMs, they have few parameters. GRUs are exploited in, e.g., \cite{Hinkka} for predictive monitoring.

In \cite{khan:2019}, instead, a Memory Augmented Neural Network (MANN) is proposed. MANNs allow to learn even longer
dependencies; training is more expensive, but MANNs are suitable for very long traces, or when cycles of the same activity may lead LSTMs or GRUs to forget/ignore activities located at the beginning of the trace.


%

The approaches which most closely relate to ours are in \cite{Philipp:20,processtransformer}, which rely on a Transformer, an architecture that substitutes the recurrence by the attention mechanism \cite{vaswani2017attention}. In particular, these works adopt multi-head attention, i.e., they perform the attention operation over different parts of the sequence simultaneously, and, instead of training an encoder-decoder architecture as in \cite{vaswani2017attention}, they only rely on the decoding part. 
As for the incorporation of external domain knowledge in the form of graphs, this approach has been explored in a few works, such as \cite{graphiT,gnn-benchmark-pe,di2017eye}, aiming to enrich sequence-based examples with geometric information derived from a graph structure. An example of this is the combination of protein structures with the geometric arrangement of molecules.
The knowledge-driven approach has also been used in several works involving pre-trained transformers (e.g. BERT) in order to align the model to a specific knowledge domain by means of ontologies and knowledge-graphs \cite{kbert,pretrain}. 
We are not aware of previous works that incorporate structural positional encoding of domain knowledge into predictive process monitoring.

\section{Methodologies}
\label{methods}
In the following sections, we are going to present the base architecture of the transformer model we are using. Subsequently, we will describe the method that enables leveraging the ontology of the activities using  Structural Positional Encoding (SPE), which is based on the Laplacian eigenvalue encoding technique \cite{gnn-benchmark-pe}.

\subsection{Transformer for PPM}
\label{archi}
\begin{figure}[h]
    \centering
    \includegraphics[scale=0.65]{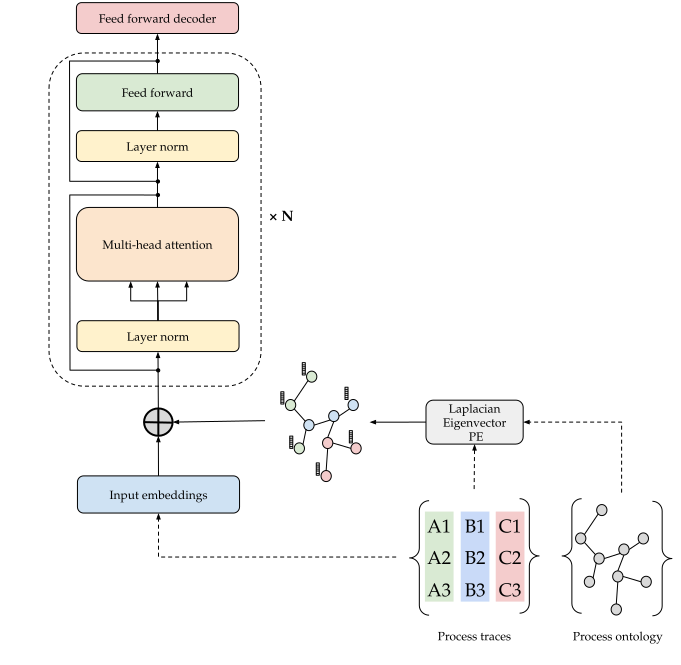}
    \caption{The model architecture. The left side of the figure represents the Transformer-decoder model, the right side shows the Structural positional encoding module.}
    \label{model-arch}
\end{figure}

Our model architecture is based on the transformer decoder (as in \cite{Philipp:20,processtransformer}). The model input is a  sequence $S = \{a_1, ..., a_i, ..., a_n\}$ (representing a trace) where every element of the sequence represents an activity coming from a set $A$ of possible activities. Figure \ref{model-arch} depicts the model structure. Below, we describe the building blocks of the model and how the inputs are processed. \newline

\noindent
\textbf{Embedding layer:} it takes $S$ as input and returns a vector of dimension $\mathbb{R}^d$ for each activity $a_i$. At this point, our samples are in the form $X \in \mathbb{R}^{n \times d}$. \newline

\noindent
\textbf{Positional encoding:} this layer is responsible for augmenting each embedding representing an activity with information about its position. In this paper, we tested two versions of positional encoding. The first version (PE henceforth), described in \cite{vaswani2017attention}, uses a sine and a cosine function to obtain a representation that respects the order of the activities within the single sequence. The second version is Structural Positional Encoding (SPE henceforth), which embeds knowledge from a graph $G$ (in our case, the graph is an ontology of the activities, and is described in section \ref{sub:spe}). Positional encoding generates a vector for each activity in a sequence that is added to the original embedding as follows:
\begin{gather*}
    X = \begin{cases}
            X + PE(X) & \text{for PE method} \\
            X + SPE(X,G) & \text{for SPE method}
        \end{cases}
\end{gather*}

\noindent
\textbf{Multi-head attention layer:} this component enables the model to selectively attend to different parts of a sequence and find long-range dependencies. To this end, the self-attention layer \cite{vaswani2017attention} generates three representation $Q,K,V$ for every activity in the sequence $S$. Subsequently, it applies the scaled dot-product attention: $$H = softmax\left(\frac{QK^T}{\sqrt{d}} \right) V$$
This operation is carried out in parallel on multiple attention heads in order to learn different types of dependencies. The resulting representations are concatenated to produce the final output.  Note that, since the task is to predict the next token in the sequence (in our application, the next activity in the trace), during the self-attention process we apply a mask so that an activity can only attend to previous ones, and ignore those that follow it. The output is finally added to the input $X$ using a skip connection. \newline

\noindent
\textbf{Layer norm:} this normalisation layer is used to mitigate vanishing or exploding gradients phenomena during the model training phase. The process ensures that the activations are centered around zero mean and unit variance making the optimization more stable \cite{layernorm}. In our implementation, we resorted to the {\em Pre-LN} configuration where the layer-norm is put inside the residual blocks which improves the model convergence \cite{preln}. \newline

\noindent
Lastly, $X$ is subsequently passed through 2 fully connected layers, where the final layer acts as a decoder, projecting the representations into a dimensionality equivalent to the number of potential activities.

\subsection{Structural Positional Encoding}
\label{sub:spe}
\begin{figure}[ht]
        \centering
        \includegraphics[width=0.53\linewidth]{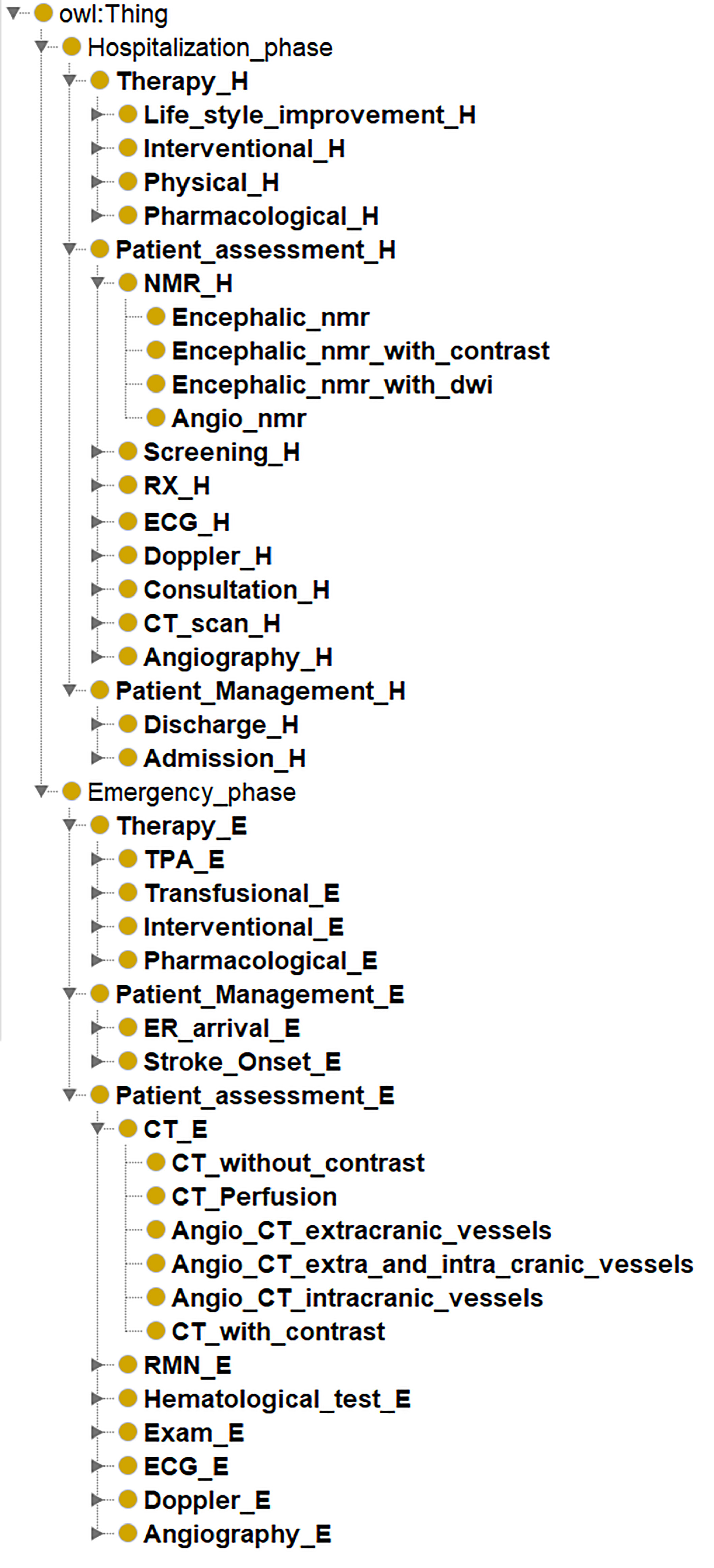}
        \caption{Excerpt of the domain ontology}
        \label{fig:onto}
\end{figure}

Structural Positional Encoding (SPE) is based on a graph representing an ontology that encodes the relations between activities. This ontology allows for the integration of  domain knowledge into the AI model, in order to improve the prediction capabilities of the system. In particular, the ontology aims at grouping together the activities involved in similar diagnostic or treatment goals. This is achieved through a taxonomic representation, where activities involved in similar goals are placed in a close "relationship"  (having a shorter path to reach one from the other). Figure \ref{fig:onto} shows an excerpt of the ontology integrated in this system. The activities and their relations  was defined in collaboration with expert neurologists, on the basis of their experience and according to the recommendations of the SPREAD guideline for stroke patients \cite{stroke_2017}.

To implement and to integrate this ontology, we coded it as a graph and we relied on the Laplacian eigenvector technique \cite{laplacian-eigenmaps}, which is employed to calculate node embeddings, yielding a vector for each action (i.e., node) that encodes its position in the graph. This ensures that nodes close to each other possess similar embeddings. During the training phase, these node representations are added to the token embeddings, before the multi-head attention layer (in a positional encoding manner). This augmentation enriches the representations of the transformer with the relational information inherent in the ontology. This enables the model to learn additional types of interactions among activities. We proceed by giving a more formal description of the operations carried out during this phase. \newline

\noindent
The graph $G=(V,E)$ representing the ontology is structured as follows: $V$ represents the nodes, one for every activity in the set $A$ (activity nodes), along with some nodes that represent the "type of activity" (activity-type nodes). $E$ captures the set of edges, connecting activity nodes to activity-type nodes. Activity-type nodes are linked together so that the graph is overall connected. The Laplacian eigenvectors are defined by the factorization of the graph Laplacian matrix: 
$$\Delta = I - D^{-\frac{1}{2}} A D^{-\frac{1}{2}} = U^T \Lambda U$$ 
where $I$ is the identity matrix, $A \in \mathbb{R}^{n \times n}$ assuming $n = |V|$ is the adjacency matrix where $A_{ij} = 1$ if node $i$ and node $j$ are connected by an edge (i.e. the graph connectivity) and $D \in R^{n \times n}$ where $D_{ii} = \sum_{i=1}^n A_{ij}$ is a diagonal matrix representing the degree of every vertex in $G$; $\Lambda$, $U$ denote the eigenvalues and eigenvectors respectively. The embedding for a node $i$ is a vector $\lambda_i \in \mathbb{R}^k$ defined as the $k$ smallest nontrivial eigenvector components of that node \cite{generalization-transformers}. Finally, to incorporate the node embedding $\lambda_i \in \mathbb{R}^k$ into the corresponding activity embedding $X_i \in \mathbb{R}^d$, we employ a fully-connected layer $\Theta \in \mathbb{R}^{k\times d}$ to ensure compatibility between the embedding sizes.

\section{Experiments}
\label{experiments}
\begin{figure}[h]
        \centering
        \includegraphics[width=0.8\linewidth]{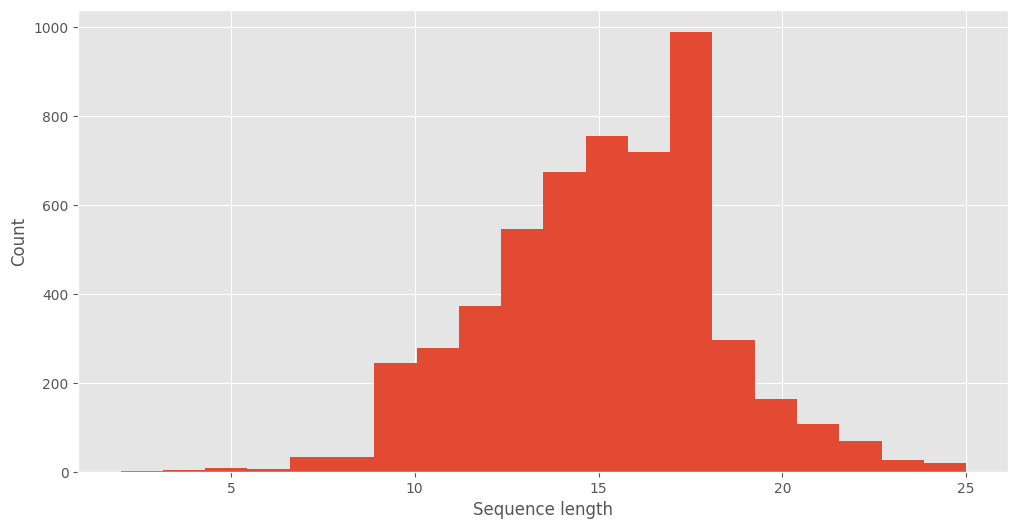}
        \caption{Sequence length histogram.}
        \label{fig:len_hist}
\end{figure}

During our experiments, we applied the model to a dataset relative to Stroke management. The dataset contains 5342 process traces with 15 activities on average and 82 possible activities (See Table \ref{tab:traces_stats} for some statistics about the traces). Given that cases have varying lengths (Figure \ref{fig:len_hist}), we applied padding to ensure that all sequences had the same number of activities (i.e. the length of the longest sequence). Moreover, we added two special tokens indicating "start of sequence" and "end of sequence" at the start and end of each case. The ontology associated to the process was parsed in order to obtain a graph containing 110 nodes and 111 edges.

The transformer model has been trained in an auto-regressive language modeling manner to predict the 'next activity' given a context (i.e., the previous activities in the sequence). The training process involves the model attempting to predict the next activity given a sequence prefix. The loss function applied is the cross-entropy loss between the model's predicted probability distribution over all possible activities and the true tokens in the training data.
Notably, during the training phase, examples containing padding tokens or "start of sequence" tokens were excluded from loss calculation. This ensures that the model ignores these examples during the calculation of gradient preventing potential degradation of the model's performance. We instead decided to keep the "end of sentence" tokens so that the model could learn to predict when a sequence of activities should end. \newline

\noindent
We performed a 80/10/10 split to the data (training, validation and test). We tuned the model hyper-parameters using the Optuna \cite{optuna} optimization framework. Table \ref{tab:grid-search} reports the search space considered and the best parameters found. \newline

\begin{table*}
\centering
\caption{Parameter search space used to find optimal hyperparameters and optimal configuration, by Optuna. Search spaces included between square brackets are discrete sets}
\begin{tabular}{lcc}
\hline
\textbf{Parameter} & \textbf{Search space}                                     & \multicolumn{1}{l}{\textbf{Best configuration}} \\ \hline
Embedding size           & {[}16, 32, 64, 128, 256{]} & 64       \\
Hidden size              & {[}16, 32, 64, 128, 256{]} & 128      \\
Number of heads          & {[}1, 2, 4, 8{]}           & 4        \\
Number of layers         & {[}1, 2, 3, 4, 5{]}        & 4        \\
Dropout rate             & 0.1 - 0.5                  & 0.216375 \\
Optimizer                & AdamW                      & AdamW    \\
Scheduler                & StepLR                     & StepLR    \\
Gamma                    & 0.85 - 0.99                & 0.989695  \\
Learning rate      & $1 \times 10^{-2}$ - $3\times10^{-2}$ & 0.002836                                        \\
Ontology embededing size & {[}8, 16, 32{]}            & 32       \\ \hline
\end{tabular}
\label{tab:grid-search}
\end{table*}

\noindent
During our experiments (see Table \ref{tab:results}) we aimed to compare how the model performed using the two different positional encoding methods (PE and SPE) and various model sizes. We also included a baseline model in which we skipped the PE phase. For each model configuration, we ran 10 fits with random training-validation-test splits and initialisation. We then reported the mean and standard deviation of accuracy-at-$k$.
The results show that there is a clear benefit in using the SPE method, which substantially improves the model performances across all model sizes. Moreover, the model metrics are quite stable and saturate at with a model size of 64, improving only marginally using 128-sized embedding. This suggests that the model does not tend to overfit.

\begin{figure}[h]
        \centering
        \includegraphics[width=0.8\linewidth]{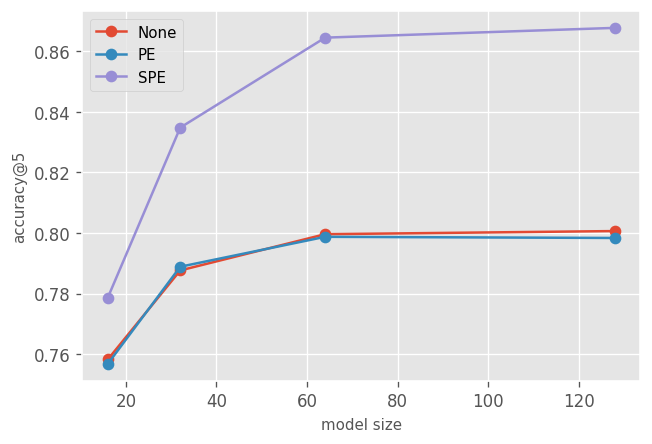}
        \caption{Model performance with different embedding sizes on testset.}
        \label{fig:plot_res}
\end{figure}

Another important finding is that the model does not seem to benefit from the classic PE method which only encodes an order to the activities (see Figure \ref{fig:plot_res}). This observation may be due to the nature of these processes, where the next activity depends more on which activities have already been performed rather than their sequential order \cite{isa}. This also hints to why the model benefits from the SPE technique: activities of a similar type are likely to be executed in close proximity. As a result, the contextual information about an activity's position within the ontology graph describing relations between activities in the process is much more informative than its position within an individual sequence.

\begin{table*}[h]
\centering
\caption{Test accuracy-at-$k$ and standard deviation on 10 random initialization and train-val-test splits. Model performance with different positional encoding methods at different embedding sizes are reported}
\label{tab:my-table}
\begin{tabular}{llccc}
\hline
\textbf{Method} & \textbf{Model size} & \multicolumn{1}{l}{\textbf{Accuracy@1}} & \multicolumn{1}{l}{\textbf{Accuracy@3}} & \multicolumn{1}{l}{\textbf{Accuracy@5}} \\ \hline
None & 16  & 45.6±0.3 & 67.1±0.3 & 75.8±0.3 \\
None & 32  & 46.9±0.2 & 69.4±0.3 & 78.8±0.3 \\
None & 64  & 47.3±0.2 & 70.4±0.2 & 80.0±0.2 \\
None & 128 & 47.2±0.1 & 70.4±0.3 & 80.1±0.2 \\ \hline
PE   & 16  & 45.2±0.3 & 66.8±0.3 & 75.7±0.2 \\
PE   & 32  & 46.9±0.2 & 69.4±0.2 & 78.9±0.2 \\
PE   & 64  & 47.1±0.1 & 70.3±0.3 & 79.9±0.2 \\
PE   & 128 & 47.0±0.2 & 70.1±0.2 & 79.8±0.2 \\ \hline
SPE  & 16  & 48.5±0.4 & 69.7±0.4 & 77.9±0.4 \\
SPE  & 32  & 52.1±0.3 & 75.3±0.4 & 83.5±0.3 \\
SPE  & 64  & 54.0±0.3 & 78.0±0.2 & 86.5±0.2 \\
SPE  & 128 & 54.3±0.3 & 78.3±0.4 & 86.8±0.3 \\ \hline
\end{tabular}

\label{tab:results}
\end{table*}

\section{Conclusions}
\label{conclu}
This paper presents an approach to predictive process monitoring, specifically applied in the challenging domain of stroke management. Leveraging the power of Transformer-based models, we have demonstrated the potential for accurate forecasting of critical information within running process traces. Our key innovation, the integration of domain-specific knowledge through Structural Positional Encoding, has been shown to increase predictive accuracy.

The initial experimental results presented in this study are encouraging, pointing towards the effectiveness of our proposed approach. These findings suggest that predictive process monitoring, using transformer-based models and domain-specific ontological knowledge, holds significant potential in providing valuable decision support in complex medical scenarios.

In the future, further research and experimentation will be crucial to validate and refine our approach. In particular, we would like to experiment with different kinds of embedding techniques for the graph nodes, and explore this methodology using different datasets, where we can explore in depth the temporal aspects.

\begin{table}[h]
\centering
\caption{Datasets traces length statistics measured in number of activities}
\label{tab:traces_stats}
\begin{tabular}{lc}
\hline
\multicolumn{2}{c}{\textbf{Traces length statistics}} \\ \hline
Mean                           & 15            \\
Standard Deviation             & 3             \\
Minimum                        & 2             \\
Maximum                        & 25            \\ \hline
\end{tabular}
\end{table}

\newpage

\subsubsection*{Acknowledgments}
We would like to thank Chameleon Cloud \cite{chameleon} for providing the essential computational resources that facilitated the execution of our experiments and model training. \newline
\noindent
Christopher Irwin and Marco Dossena are a PhD students enrolled in the National PhD in Artificial Intelligence, XXXVIII cycle, course on Health and life sciences, organized by Università Campus Bio-Medico di Roma.

\subsubsection*{Code availability}
The source code of the transformer model and the implementation of the SPE module are available at: \href{https://github.com/christopher-irw/proformer\_ce}{github.com/christopher-irw/proformer\_ce}. The repository contains the necessary code to train the transformer model on the BPI 2012 challenge \cite{bpi12}. We also provide an example of a very simple ontology of the actions present in the dataset in order to make the SPE method applicable.

\bibliography{sn-bibliography}

\end{document}